\documentclass[conference]{IEEEtran}
\IEEEoverridecommandlockouts
\usepackage{cite}
\usepackage{amsmath,amssymb,amsfonts}
\usepackage{algorithmic}
\usepackage{graphicx}
\usepackage{textcomp}
\usepackage{xcolor}
\usepackage{comment}
\usepackage{pifont}
\usepackage{booktabs} 
\usepackage{multirow} 
\usepackage[utf8]{inputenc} 
\usepackage{graphicx} 
\usepackage{caption}
\usepackage{subcaption}
\usepackage{authblk}
\usepackage{hyperref}

\def\BibTeX{{\rm B\kern-.05em{\sc i\kern-.025em b}\kern-.08em
    T\kern-.1667em\lower.7ex\hbox{E}\kern-.125emX}}
\begin{document}

\title{Dynamic Token Selective Transformer for Aerial-Ground Person Re-Identification}

\author{
  Yuhai Wang$^{1*}$, Maryam Pishgar$^1$ \\ 
  {\small $^{1}$University of Southern California} \\ 
  {\small $^{*}$Corresponding author}\\
  {\small \href{https://yuhaiw.github.io/DTS-AGPReID/}{\textbf{https://yuhaiw.github.io/DTS-AGPReID/}}} 
}

\maketitle

\begin{abstract}
Aerial-Ground Person Re-identification (AGPReID) holds significant practical value but faces unique challenges due to pronounced variations in viewing angles, lighting conditions, and background interference. Traditional methods, often involving a global analysis of the entire image, frequently lead to inefficiencies and susceptibility to irrelevant data. In this paper, we propose a novel Dynamic Token Selective Transformer (DTST) tailored for AGPReID, which dynamically selects pivotal tokens to concentrate on pertinent regions. 
Specifically, we segment the input image into multiple tokens, with each token representing a unique region or feature within the image. Using a Top-k strategy, we extract the k most significant tokens that contain vital information essential for identity recognition. Subsequently, an attention mechanism is employed to discern interrelations among diverse tokens, thereby enhancing the representation of identity features. Extensive experiments on benchmark datasets showcases the superiority of our method over existing works. Notably, on the CARGO dataset, our proposed method gains 1.18\% mAP improvements when compared to the second place. 
In addition, we comprehensively analyze the impact of different numbers of tokens, token insertion positions, and numbers of heads on model performance.

\end{abstract}


\section{Introduction}
\label{sec:intro}
Person Re-identification (ReID) is crucial for surveillance and tracking, identifying individuals across camera views. Advances in deep learning have improved feature extraction and matching accuracy~\cite{sun2019dissecting, chen2022self, tang2023spike, zhang2023pose, chen2024region}. However, most methods rely on global image features, making them vulnerable to background noise and irrelevant regions, particularly in cases of occlusion or complex backgrounds. This limits their effectiveness in diverse real-world scenarios with cross-camera variations and environmental inconsistencies~\cite{lee2023camera, li2021weperson, zhang2022uncertainty}.


To address these challenges, recent studies have emphasized the importance of more targeted and efficient feature extraction approaches. For instance, Zhang et al. \cite{zhang2024separable} propose a separable attention mechanism to focus on discriminative regions while suppressing irrelevant background features. Tang et al. \cite{tang2023ac2as} introduce adaptive context-aware selection to dynamically enhance feature representations under complex conditions.  
\begin{figure}[htbp]
    \centering
    \includegraphics[width=0.48\textwidth]{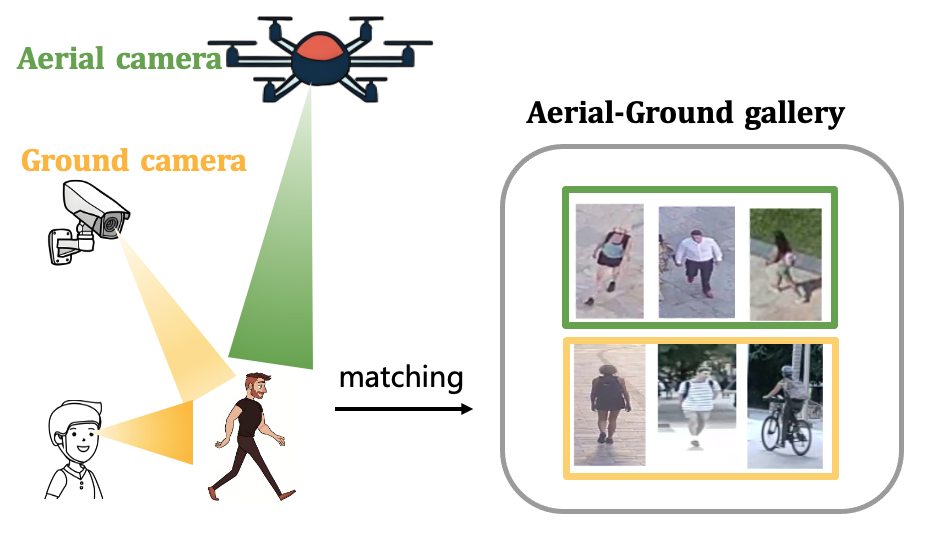}
    \caption{A straightforward description of Aerial-Ground Person Re-identification (AGPReID) involves the utilization of an aerial-ground mixed camera network, enabling matching across aerial-aerial, ground-ground, and aerial-ground scenarios. Thus, it presents greater challenges and practical applications compared to traditional single-camera person ReID methods.}
    \label{fig:intro} 
\end{figure}
Similarly, Qiu et al. \cite{qiu2024salient} develop a salient feature extraction framework that prioritizes key object parts even in scenarios involving significant occlusion. These advancements show promising progress in overcoming the limitations of the reliance on global feature in View-homogeneous person ReID. However, when applied to Aerial-Ground Person Re-identification (AGPReID) tasks (View-heterogeneous person ReID), which are valuable in real-world scenarios for addressing complex aerial-to-ground matching challenges and encompassing diverse camera perspectives \cite{nguyen2023aerial}, these methods often fall short. Fig.~\ref{fig:intro} demonstrates the AGPReID problem. This discrepancy may stem from the scale diversity and redundancy characteristics observed in large-area observational scenarios, leading to notable appearance differences for the same individual across various cameras. Therefore, there is an urgent need to develop innovative strategies that effectively address these specific challenges in AGPReID. 


\begin{figure*}[htbp]
    \centering
    \includegraphics[width=0.8\textwidth]{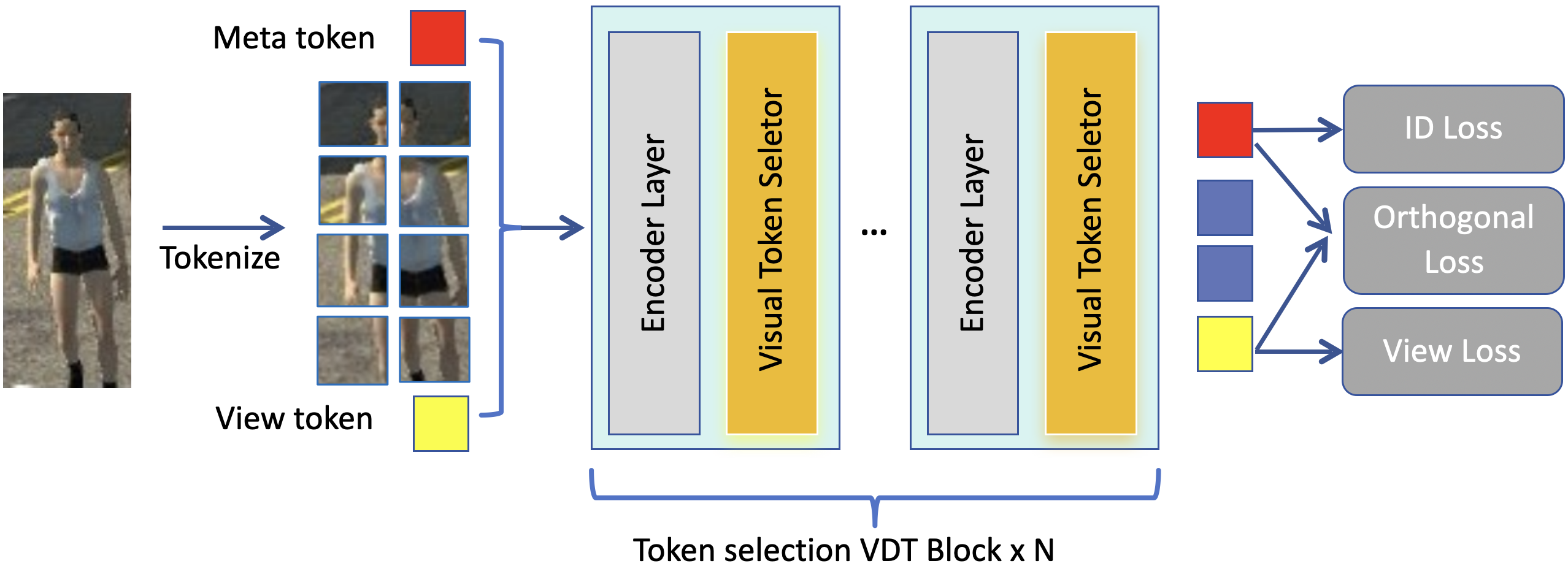}
    \caption{
Illustration of the proposed Dynamic Token Selective Transformer (DTST) framework. The framework incorporates $N$ Token Selection  view-decoupled transformer (VDT) blocks, where each block consists of an encoder layer and a visual token selector. The loss function is designed to account for both view-related and view-unrelated features, while an orthogonal loss ensures that these features remain independent from each other, further enhancing feature disentanglement and robustness.
}

    \label{fig:Tk} 
\end{figure*}

\begin{figure*}[htbp]
    \centering
    \includegraphics[width=\textwidth]{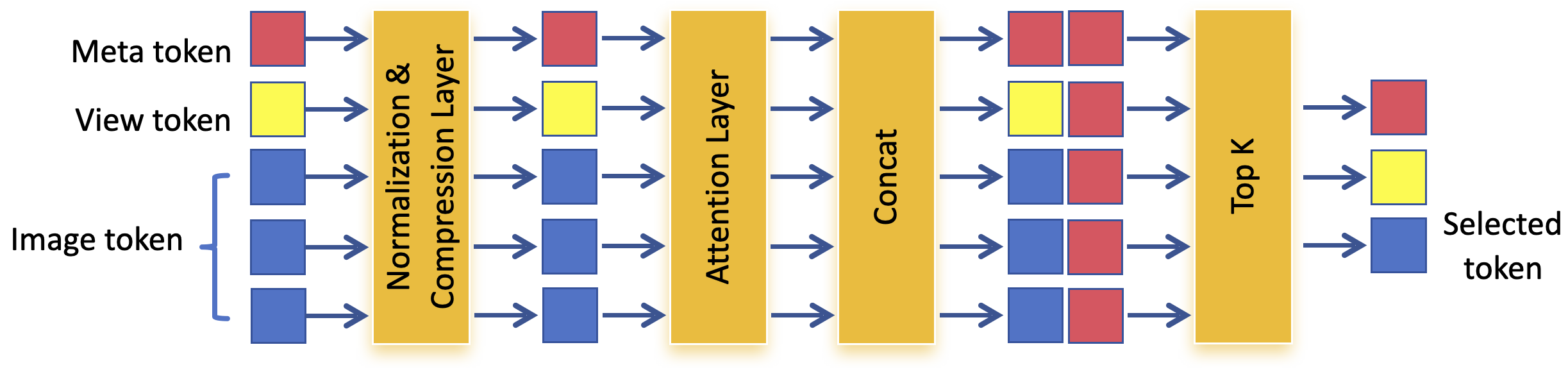}
    \caption{The Illustration of  Visual Token Selector (VTS). The process involves selecting the Top-K informative tokens from the original token set to be used in the subsequent feature aggregation.}
    \label{fig:Ts} 
\end{figure*}

To this end, we propose a Dynamic Token Selective Transformer (DTST) that enhances identity representation by focusing on the most critical spatial features. Our DTST module contains two steps: First, a Predictor Local-Global network computes relevance scores for each token, integrating local and global spatial semantics using multi-head attention. Second, a Perturbation-Based Top-K Selector chooses the most relevant tokens based on the predicted scores, ensuring robustness by adding noise perturbations. The selected tokens are combined with a global class token, enabling efficient and compact representation while reducing computational overhead. Extensive experiments validate our method's state-of-the-art performance on AGPReID tasks, showcasing its robustness in handling occlusions, complex backgrounds, and viewpoint variations.

Our main contributions are as follows.
\begin{itemize}
\item We propose a Top-k Token Selective Transformer for AGPReID, to better model identity representation spatially. We further comprehensively study the impact of the insertion position and the number of tokens selected on the model's performance.
\item To eliminate the interference of irrelevant tokens, our method adaptively selects the most critical tokens based on the top-k selective mechanism, making the long-range modeling more effective and compact.
\item Extensive experiments on various datasets demonstrate that our proposed model achieves state-of-the-art performance on AGPReID tasks.
\end{itemize}

\section{Related Work}
\subsection{Person Re-identification}
Person re-identification (ReID) is essential for retrieving images of the same individual across different camera views. It can be categorized into view-homogeneous and view-heterogeneous ReID. View-homogeneous ReID pertains to scenarios with a single camera type, such as ground-only or aerial-only networks, while view-heterogeneous ReID such as Aerial-Ground Person ReID (AGPReID), deals with networks featuring diverse camera perspectives.
In terms of view-homogeneous ReID, ground-only camera networks have received more attention compared to aerial-only networks. For example, some ground-only datasets are well established such as Market1501\cite{zheng2015scalable} and MSMT17\cite{wei2018person}. As a consequence, a multitude of methods have been proposed, such as hand-crafted feature-based, CNN-based, and transformer-based approaches, facilitating the development of ReID. However, these methods overlook the significant view differences between aerial and ground cameras, leading to poor performance faced with diverse view-point scenarios. Fortunately, view-heterogeneous ReID can address this issue. Recently, researchers in~\cite{nguyen2023aerial} propose the AG-ReID dataset, which includes identity and attribute labels, and put forward an attribute-guided model. Another work extends this by introducing the CARGO dataset with multiple matching scenarios and proposes a view-decoupled transformer (VDT) that decouples view-related features using hierarchical separation and orthogonal loss, improving performance and reducing reliance on extensive attribute labeling \cite{zhang2024view}. However, this approach does not dynamically select key tokens related to the target object, fails to reduce redundant computation, and lacks enhanced model capability to focus specifically on critical regions of interest.

\subsection{Token Selection in Vision Transformers}
Token selection is crucial for addressing redundancy issues in transformer-based vision models, particularly in tasks involving dense visual data. Despite their success, transformers often suffer from computational inefficiencies due to the need to process numerous redundant tokens. Token selection methods can effectively mitigate this issue by focusing on only the most informative tokens for further processing.
For example, STTS \cite{wang2022efficient}, as a representative work, utilizes token selection to enhance computational efficiency by dynamically reducing the number of tokens processed at each transformer layer. These approaches have demonstrated substantial reductions in computation while maintaining performance. To address the challenge of differentiability in token selection, a perturbed maximum strategy is introduced \cite{berthet2020learning}, enabling top-K selection to be differentiable, thereby facilitating end-to-end training. Building on the principles of differentiable top-K selection \cite{liu2022ts2}, we develop a lightweight token selection module specifically designed to enhance temporal-spatial modeling in our view-decoupled transformer. By selecting only the most informative tokens, this module reduces redundancy and improves both efficiency and performance, especially in modeling visual data across multiple viewpoints.

\section{Method}
\subsection{Formulation}
Aerial-Ground Person ReID aims to match images from ground- or aerial-only camera networks. In a training dataset $\mathcal{D}^{tr}=\left\{ \left( x_i,y_i,v_i \right) \right\} _{i=1}^{|\mathcal{D}^{tr}|}$, each instance consists of an image $x_i$ depicting a person, along with identity label $y_i$ and view label $v_i$. The view label $v_i \in \left\{v^a, v^g\right\}$ is determined by the known camera labels in $\mathcal{D}$, distinguishing between aerial ($v^a$) and ground ($v^g$) views. A substantial distinction in views between $v^a$ and $v^g$ results in a biased feature space, characterized by low intra-identity similarity and high inter-identity dissimilarity.

\subsection{Overview}

As illustrated in Fig.\ref{fig:Tk}, we propose a token enhanced framework based on the View-Decoupled Transformer (VDT) to tackle the view discrepancy challenge in AGPReID. 
Input images that include both aerial (\( v_a \)) and ground (\( v_g \)) views are tokenized into a sequence of tokens. To encompass both global and view-specific details, meta tokens and view tokens are added to these image tokens before they are inputted into our VDT.

Comprising \( N \) blocks, the VDT framework initiates each block with a conventional self-attention encoding process, succeeded by a subtraction operation between meta and view tokens to explicitly disentangle view-specific characteristics from the overarching ones. This facilitates a distinct segregation of features influenced by diverse viewpoints.

Subsequently, the updated meta and view tokens produced by the VDT are supervised by identity and view classifiers. To enforce the independence of meta and view tokens, we introduce an orthogonal loss, facilitating the successful separation of view-based and view-agnostic attributes. To select the most critical tokens, a \textbf{visual token selector} module is proposed to enhance the identity representation, with further elaboration provided in subsequent sections.





We introduce the Visual Token Selector (VTS), as shown in Fig.~\ref{fig:Ts},  designed to dynamically refine the token representation by selecting the most informative tokens for subsequent analysis. This module aims to reduce redundancy and enhance the model's ability to focus on critical regions, thereby optimizing computational efficiency while preserving feature quality. The VTS mechanism can be understood as a dynamic token selection process that leverages attention scores to determine the importance of each token.

For a sequence of tokens \(\{t_i\}_{i=1}^{M}\), where \(M\) is the number of tokens, the VTS computes importance scores for each token \(s_i\) using a lightweight attention mechanism. The score \(s_i\) is obtained as:

\[
s_i = \text{softmax}\left(\frac{t_i^\top W_q W_k^\top t_i}{\sqrt{d}}\right),
\]

where \(t_i\) is the \(i\)-th token, \(W_q\) and \(W_k\) are learnable matrices representing query and key transformations, and \(d\) is the dimensionality of the tokens. The softmax function normalizes the scores to ensure they sum to 1, thus creating a probabilistic distribution over the tokens.

These tokens are then ranked based on their importance scores, and we select the top-\(K\) tokens with the highest scores, where \(K < M\) is a hyperparameter that controls the number of tokens retained. Mathematically, this selection can be represented as:

\[
\{t_i^{\text{selected}}\} = \text{TopK}(\{s_i\}_{i=1}^{M}),
\]

where \(\text{TopK}(\cdot)\) returns the indices corresponding to the top-\(K\) scores. The retained tokens, \(\{t_i^{\text{selected}}\}\), are then passed to the subsequent layers or directly to the final classification head.

To ensure that the VTS can be used in an end-to-end training fashion, we adopt a differentiable approach for the token selection. Specifically, we use a continuous relaxation of the \(\text{TopK}\) function by employing a Gumbel-Softmax trick:

\[
\hat{s}_i = \frac{\exp((s_i + g_i)/\tau)}{\sum_{j=1}^{M} \exp((s_j + g_j)/\tau)},
\]

where \(g_i\) are Gumbel noise samples and \(\tau\) is the temperature parameter that controls the smoothness of the approximation. This differentiable approximation allows the selection of tokens to be included in backpropagation, facilitating end-to-end optimization.

By incorporating the Visual Token Selector, we achieve several key benefits:
\begin{table*}[t]
	\centering
        \footnotesize
        \renewcommand{\arraystretch}{0.4}
	\caption{THE DETAILED SUMMARY OF THE DATASET PROPERTIES INVOLVED IN THIS PAPER, INCLUDING AG-ReID and CARGO.}
	\label{datasets}
	\renewcommand{\arraystretch}{1.1}
	\resizebox{0.6\textwidth}{!}{%
		\begin{tabular}{c|c|c|c|c|c}
			\hline
			Dataset & Data &\#PersonID & \#Camera & \#Image & \#Height \\ \hline
			AG-ReID~\cite{10403853} & Real & 388 & 2 (1A+1G) & 21,893 & $15\sim45m$ \\
			CARGO~\cite{zhang2024view} & Synthetic & 5,000 & 13 (5A+8G) & 108,563 & $5\sim75m$ \\ \hline
		\end{tabular}
	}
\end{table*}
\begin{table*}[t]
    \centering
    \caption{Performance comparison of the mainstream methods under four settings of the proposed CARGO dataset. ``ALL'' denotes the overall retrieval performance of each method. ``G$\leftrightarrow$G,'' ``A$\leftrightarrow$A,'' and ``A$\leftrightarrow$G'' represent the performance of each model in several specific retrieval patterns. Rank1, mAP, and mINP are reported (\%). The best performance is shown in \textbf{bold}.}
    \label{tab:comparison_results}
    \resizebox{\textwidth}{!}{%
    \begin{tabular}{@{}lcccccccccccc@{}}
    \toprule
    \multirow{2}{*}{Method} & \multicolumn{3}{c}{Protocol 1: ALL} & \multicolumn{3}{c}{Protocol 2: G$\leftrightarrow$G} & \multicolumn{3}{c}{Protocol 3: A$\leftrightarrow$A} & \multicolumn{3}{c}{Protocol 4: A$\leftrightarrow$G} \\ 
    \cmidrule(lr){2-4} \cmidrule(lr){5-7} \cmidrule(lr){8-10} \cmidrule(lr){11-13}
    & Rank1 & mAP & mINP & Rank1 & mAP & mINP & Rank1 & mAP & mINP & Rank1 & mAP & mINP \\ \midrule
    SBS \cite{SBS2021} & 50.32 & 43.09 & 29.76 & 72.31 & 62.99 & 48.24 & 67.50 & 49.73 & 29.32 & 31.25 & 29.00 & 18.71 \\
    PCB [37] & 51.00 & 44.50 & 32.20 & 74.10 & 67.60 & 55.10 & 55.00 & 44.60 & 27.00 & 34.40 & 30.40 & 20.10 \\
    BoT \cite{luo2019bag} & 54.81 & 46.49 & 32.40 & 77.68 & 66.47 & 51.34 & 65.00 & 49.79 & 29.82 & 36.25 & 32.56 & 21.46 \\
    MGN \cite{wang2018learning} & 54.81 & 49.08 & 36.52 & \textbf{83.93} & 71.05 & 55.20 & 65.00 & 52.96 & 36.78 & 31.87 & 33.47 & 24.64 \\
    VV [40, 41] & 45.83 & 38.84 & 39.57 & 72.31 & 62.99 & 48.24 & 67.50 & 49.73 & 29.32 & 31.25 & 29.00 & 18.71 \\
    AGW \cite{AGW2020} & 60.26 & 53.44 & 40.22 & 81.25 & 71.66 & 58.09 & 67.50 & 56.48 & 40.40 & 43.57 & 40.90 & 29.39 \\
    ViT \cite{dosovitskiy2020image} & 61.54 & 53.54 & 39.62 & 82.14 & 71.34 & 57.55 & 80.00 & \textbf{64.47} & \textbf{47.07} & 43.13 & 40.11 & 28.20 \\
    VDT \cite{zhang2024view} & 62.82  & 54.22 & 39.92 & 79.46 & 70.60 & 57.89  & \textbf{82.50}  & 64.06 & 44.67 & 47.50 & 42.21 & 29.03 \\
    \textbf{DTST (Ours)}  & \textbf{64.42} & \textbf{55.73} & \textbf{41.92} & 78.57 & \textbf{72.40} & \textbf{62.10} & 80.00 & 63.31 & 44.67 & \textbf{50.63} & \textbf{43.39} & 
    \textbf{29.46}\\

    \bottomrule
    \end{tabular}%
    }
    \label{tab:cargo_comparison}
\end{table*}
\begin{itemize}
    \item \textbf{Reduce redundancy}: By selecting only the most informative tokens, we minimize the amount of redundant information processed by the model.
    \item \textbf{Enhance discriminability}: The model can focus on the most critical aspects of the input, leading to improved performance on tasks requiring fine-grained feature analysis.
    \item \textbf{Improve computational efficiency}: Reducing the number of tokens processed helps in reducing the overall computational cost, making the model more efficient for both training and inference.
\end{itemize}

\section{Experiments}
\subsection{Experiment settings}

\textbf{Datasets.}
We conduct experiments on the CARGO~\cite{zhang2024view} dataset and AG-ReID dataset~\cite{10403853}. Compared to AG-ReID, the CARGO dataset offers a larger scale, greater diversity, and is the first large-scale synthetic dataset for AGPReID. Table \ref{datasets} summarizes both datasets. For CARGO, 51,451 images with 2,500 IDs are used for training, and 51,024 images with 2,500 IDs for testing. Four evaluation protocols (ALL, A$\leftrightarrow$A, G$\leftrightarrow$G, and A$\leftrightarrow$G) assess model performance, with A$\leftrightarrow$A and G$\leftrightarrow$G testing aerial and ground data separately, and A$\leftrightarrow$G using cross-view retrieval. The training set is consistent across all protocols.

For AG-ReID, 11,554 images with 199 IDs are used for training, and 12,464 images with 189 IDs for testing. Two protocols, A$\rightarrow$G and G$\rightarrow$A, evaluate cross-view retrieval, with the former testing 1,701 aerial queries against 3,331 ground galleries, and the latter 962 ground queries against 7,204 aerial galleries.

\textbf{Evaluation Metrics.}
Following the common setting, we utilize three metrics to evaluate our model: the cumulative matching characteristic at Rank1, mean Average Precision (mAP), and mean Inverse Negative Penalty (mINP).
\subsection{Implementation Details}
Our model is implemented using the PyTorch framework, with experiments conducted on an NVIDIA 4090 GPU. We use the View-decoupled Transformer (VDT) as the baseline, which includes 12 transformer encoder blocks based on the ViT-Base architecture, pre-trained on ImageNet with a patch size and stride of 16×16. Input images are resized to 256×128 during preprocessing. The training process employs the Stochastic Gradient Descent (SGD) optimizer with a cosine learning rate decay, starting at $8 \times 10^{-3}$ and reducing to $1.6 \times 10^{-6}$ over 120 epochs. The batch size is set to 128, comprising 32 identities with four images per identity. Our token selector module features a two-head transformer encoder that selects the top two rated tokens, integrated after the final transformer encoder block for enhanced performance.

\subsection{Comparisons with State-of-the-art Methods}
We evaluate our proposed DTST against state-of-the-art methods on the CARGO and AG-ReID datasets, comprising CNN-based approaches (BoT \cite{luo2019bag}, SBS \cite{SBS2021}, MGN \cite{wang2018learning}, AGW \cite{AGW2020}) and transformer-based methods (ViT \cite{dosovitskiy2020image}, VDT \cite{zhang2024view}).\\
\textbf{Performance on CARGO.}
Table \ref{tab:comparison_results} shows the results of our proposed DTST and other competitive methods on the CARGO dataset. The proposed DTST achieves state-of-the-art performance. For example, DTST surpasses the mAP/Rank-1/mINP of the baseline by 1.18\%/3.13\%/0.43\% on the aerial-to-ground (A$\leftrightarrow$G) protocol of CARGO. Besides, DTST also brings different degree of benefits to other CARGO protocols. Specifically, our proposed DTST exceeds VDT on mAP/Rank-1/mINP by 1.51\%/1.60\%/2.00\% on the ALL of AG-ReID. Demonstrating the effectiveness of the Dynamic Token Selective Transformer in mitigating view bias and improving identity representation. Previous view-homogeneous ReID methods show significant performance degradation under the view-heterogeneous AGPReID protocols, especially in cases of considerable view variation. This decline underscores how view bias hampers the consistency of identity features across views. Unlike existing methods that overlook this key challenge and struggle to generalize in heterogeneous scenarios, our approach adaptively selects the most critical tokens using a top-k selective mechanism. This token selection not only maintains accuracy but even enhances it, resulting in more effective and compact long-range modeling.\\
\textbf{Performance on AG-ReID.}
To further demonstrate the performance of our model, we also carry out similar experiments on the AG-ReID dataset. The outcomes are detailed in Table \ref{tab:performance-agreid}. As depicted in Table \ref{tab:performance-agreid}, we compare two challenging protocols: A$\rightarrow$G and  G$\rightarrow$A. 
It is noteworthy that VDT serves as a strong baseline. However, our proposed method, DTST, demonstrates a significant enhancement, outperforming VDT by 0.57\% for the A$\rightarrow$G Rank-1 protocol and 1.04\% for the G$\rightarrow$A Rank-1 protocol. 
This consistent improvement suggests that the superior performance of DTST does not stem from a robust baseline VDT but from the proposed method itself.

\begin{table}[t]
	\centering
	\caption{Quantitative evaluation of the mainstream methods under two settings of AG-ReID dataset. ``A$\leftrightarrow$G'', and ``G$\leftrightarrow$A'' represent the performance in two specific patterns. Rank1, mAP, and mINP are reported (\%). Best marked in \textbf{bold}.}
	\renewcommand{\arraystretch}{1.2}
	\resizebox{0.47\textwidth}{!}{%
		\begin{tabular}{c|ccc|ccc}
			\hline
			\multirow{2}{*}{Method} & \multicolumn{3}{c|}{Protocol 1: A$\rightarrow$G} & \multicolumn{3}{c}{Protocol 2: G$\rightarrow$A} \\ \cline{2-7} 
			& Rank1 & mAP & mINP & Rank1 & mAP & mINP \\ \hline
			SBS \cite{SBS2021} & 73.54 & 59.77 & - & 73.70 & 62.27 & - \\
			BoT \cite{luo2019bag} & 70.01 & 55.47 & - & 71.20 & 58.83 & - \\
			OSNet \cite{zhou2021learning} & 72.59 & 58.32 & - & 74.22 & 60.99 & -  \\
			\hline
			ViT \cite{dosovitskiy2020image} & 81.28 & 72.38 & - & 82.64 & 73.35 & - \\ 
			VDT \cite{zhang2024view} & 82.91 & 74.44 & \textbf{51.06} & 83.68 & 75.96 & 49.39\\
            \textbf{DTST (ours)} & \textbf{83.48} & \textbf{74.51} & 49.86 & \textbf{84.72} & \textbf{76.05} & \textbf{50.04}\\
            \hline
		\end{tabular}%
	}\label{tab:performance-agreid}
\end{table}

\begin{table}[t]
	\caption{Ablation study of model key designs on CARGO dataset. Rank1, mMAP, and mINP are reported(\%). Best in \textbf{blod}.}
	\centering
	\label{table:adlation-tk}
	\resizebox{0.47\textwidth}{!}{
		\begin{tabular}{lccccc}
			\hline
			\multirow{2}{*}{Method} & \multirow{2}{*}{Visual Token Selector}  & \multicolumn{3}{c}{Protocol: A$\leftrightarrow$G}  \\
			& & Rank1 & mAP & mINP  \\
			\hline
            model-a & \ding{56} & 45.00 & 42.05 & \textbf{30.26}\\
            model-b (\textbf{Ours}) & \ding{52} & \textbf{50.63} & \textbf{43.39} & 29.46 \\ 
			\hline
	\end{tabular}}
\end{table}

\begin{table}[t]
	\caption{Ablation study on the number of attention heads, token quantities, and token positions using the CARGO dataset. ``Head-Num.'' signifies the quantity of attention heads, ``T-Num.'' demotes the number of token, and ``T-Position.'' indicates the specific position where each token is locate. Performance is assessed through Rank1, mAP, and mINP(\%), with the best results highlighted in \textbf{blod}.}
	\centering
	\label{table:adlation-2}
	\resizebox{0.47\textwidth}{!}{
		\begin{tabular}{cccccccccc}
			\hline
			\multirow{2}{*}{Method} & \multirow{2}{*}{Head-Num.} & \multirow{2}{*}{T-Num.} & \multirow{2}{*}{T-Position.} &
            \multicolumn{3}{c}{Protocol: A$\leftrightarrow$G}  \\
			& & & & Rank1 & mAP & mINP  \\
			\hline
            model-1 & 8 & 2 & last layer & 46.25 &42.56 &\textbf{30.16} \\
            model-2 & 8 & 3 & last layer &45.00 &41.28 &28.83\\
            model-3 & 8 & 3 & second to last layer & 46.88 &41.04 &28.12\\
            model-4 & 8 & 32 & second to last layer  & 40.00 &36.58 &24.73\\
            model-5 & 4 & 2 & last layer& 46.88 &42.46 &29.79\\
            model-6 (\textbf{Ours}) & 2 &2& last layer&\textbf{50.63} & \textbf{43.39} & 29.46\\ 
			\hline
	\end{tabular}}
\end{table} 
       
\subsection{Ablation Study}
In this section, we provide ablation study to investigate several key components of our DTST. We also delved into the number of attention heads, token quantities, and token positions. Notably, all ablation experiments are conducted on the on the CARGO dataset.\\
\textbf{Effects of Visual Token Selector (VTS).} We first explore the effectiveness with placing the Visual Token Selector before the final layer of the View-decoupled Transformer. In this setup, all other settings,such as the number of attention heads and selected tokens, remain constant.
Table \ref{table:adlation-tk} shows the results, where model-a lacks a visual token selector, whereas model-b incorporates one. 
From the Table, we can observe a 5.63\% improvement in rank-1 accuracy and 1.34\% increase in mAP accuracy under Protocol A$\rightarrow$G, which indicates that the token selection strategy effectively filters out tokens with discriminative features and eliminates identity-irrelevant tokens, thereby enhancing better identity representation.\\
\textbf{Number of Heads.} We also evaluate the performance of VTS with different numbers of heads, specifically 2, 4, and 8 heads in Table \ref{table:adlation-2}. Interestingly, using more heads results in a decrease in accuracy. Specifically, when increasing the number of heads from 2 to 4, there is a 3.64\% decline in rank-1 accuracy and 0.93\% drop in mAP. This suggests that a higher number of heads may dilute the model's ability to focus on critical identity features, potentially introducing noise and decreasing overall model performance. One underlying reason may be model over-fitting, as a greater number of heads could increase the model’s complexity without corresponding improvements in performance. Another potential explanation might be that more heads may dilute the importance of the most vital tokens, leading to less effective feature aggregation.\\
\textbf{Number of Tokens Selected.} Keeping other variables constant, we analyze the impact of different numbers of token selections on model performance in Table \ref{table:adlation-2}. 
We vary the number of tokens to 2, 3, 5.
The findings reveal that selecting 2 or 3 tokens yields superior results across all evaluation metrics, i.e. Rank-1 accuracy, mAP, and mINP. 
Specifically, we increase the number of selected tokens beyond 3, but the performance fails to show any improvement, indicating that opting for fewer but more critical tokens enables the model to concentrate better on pivotal identity features. In contrast, selecting more tokens may introduce irrelevant information, thereby compromising overall accuracy. When our method is applied in the same setup, choosing 3 tokens compared to 2 tokens results in a decrease of 1.25\% in rank-1 accuracy, 1.28\% in mAP, and 1.33\% in mINP, highlighting the trade-off between token quantity and model's focus on essential features.\\
\textbf{Token positions.} 
The insertion position of VST, whether in the last or second-to-last layer, also affects model performance, as shown in Table \ref{table:adlation-2}.
When the fixed number of heads is 8 and the number of tokens is 3, model-3 achieves a higher Rank-1 accuracy at 46.88\%, but both mAP and mINP decrease. 
The reason behind this could be that tokens in shallow layers contain more detailed information, while tokens in deeper layers extract higher-level semantic information. As a result, the information within each token becomes more refined, leading to a higher compressibility ratio.

\section{Conclusion and Future Work}
In this paper, we investigate the relationships between tokens in transformers and propose a dynamic token selective transformer specifically for the AGReID task.
Experiments demonstrate that incorporating token selection can effectively reduce token redundancy, enhance the importance of discriminative tokens, and consequently achieve state-of-the-art results. 
Furthermore, we investigated the impact of different implementation details, the number of tokens, and the position of token insertion on model performance, providing a comprehensive understanding of the influence of token selection on AGReID.
Token selection is a general technique, and we will explore its application in other tasks. While our work focuses on token-level selection, recent studies demonstrate the potential of pixel-level operations\cite{nguyen2024image}, showing effectiveness in tasks like object classification, masked autoencoding, and image generation. Inspired by this, we aim to integrate token and pixel selection to enhance the efficiency and performance of vision models.

\bibliographystyle{IEEEbib}
\bibliography{icme2025_template_anonymized}

\end{document}